# Blind Modulation Classification based on MLP and PNN

Harish Chandra Dubey[1], Nandita[2], and Ashutosh Kumar Tiwari[3]

*Abstract*-- In this work, a pattern recognition system is investigated for blind automatic classification of digitally modulated communication signals. The proposed technique is able to discriminate the type of modulation scheme which is eventually used for demodulation followed by information extraction. The proposed system is composed of two sub-systems namely feature extraction sub-system (FESS) and classifier sub-system (CSS). The FESS consists of continuous wavelet transform (CWT) for feature generation and principal component analysis (PCA) for selection of the feature subset which is rich in discriminatory information. The CSS uses the selected features to accurately classify the modulation class of the received signal. The proposed technique uses probabilistic neural network (PNN) and multi-layer perceptron forward neural network (MLPFN) for comparative study of their recognition ability. PNN have been found to perform better in terms of classification accuracy as well as testing and training time than MLPFN. The proposed approach is robust to presence of phase offset and additive Gaussian noise.

*Index Terms*-- Blind modulation recognition, continuous wavelet transform, electronic surveillance, modulation classification, multi-resolution analysis, probabilistic neural network, software defined radios.

## I. INTRODUCTION

The importance of blind signal interception cannot be overlooked in domain of wireless communication systems. Effective, accurate and robust automatic blind modulation classification systems have gained importance in light of development of 3G and 4G systems. They gained popularity in both military as well as civilian applications. The military application includes electronic support measures, spectrum surveillance, interference identification and threat evaluation. The civilian applications include software defined radios (SDRs), cognitive radios etc. [1]-[2]. In this work, a novel technique based on probabilistic neural network (PNN) has been suggested for blind automatic classification of digitally modulated communication signals. The continuous wavelet transform (CWT) have become popular for feature generation in context of modulation classification. The artificial neural networks have emerging application in classification of digitally modulated signals. So, the proposed technique integrates wavelet features with PNN. Moreover the proposed PNN classifier has been compared with multilayer perceptron feed forward neural network (MLPFN) described in [3]. The feature extraction and classifier sub-systems are supposed to be independent but there is no specific study regarding this issue in literature known to author.

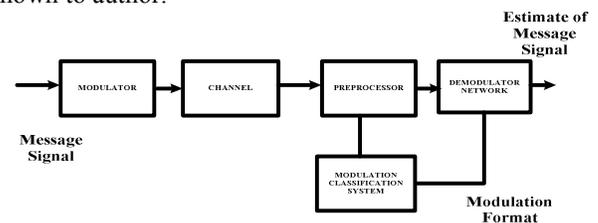

Fig.1. An intelligent communication system

The block diagram of an intelligent communication system is shown in Fig.1. It shows that modulation classification is an intermediate step between signal detection and information extraction. Such a system is expected to have:

- High speed so that signals can be identified in least time
- Flexibility so that the system can be put to use if some additional modulation class is included.
- Robust so that the change in design assumptions will not affect the performance of the classifier significantly.

A comprehensive study of all available techniques for modulation classification has been made in [4]. Other work in related aspects includes [4]-[12]. In spite of various research studies done it is not possible to compare the methods available in literature due to non availability of standard database of digitally modulated signals. Most the studies did incorporate additive white Gaussian noise (AWGN) channel model [4]. The proposed study incorporates the flat fading channel with phase offset and have considered QAM-16, QAM-64, ASK-4, ASK-8, PSK-2, PSK-4, PSK-8, FSK-2, FSK-4 and MSK modulation schemes. To evaluate the performance of the proposed technique extensive simulation tests has been performed under various conditions. The remainder of the paper is organized as follows. Section II describes the underlying mathematical model. Section III accounts for blind classification technique used in this work. Section IV focuses on simulation tests with varying conditions. Finally, conclusions are drawn on the basis of simulation studies in section V.

---
[1]Harish Chandra Dubey is with the Department of Electronics and Communication Engineering, Motilal Nehru National Institute of Technology(MNNIT), Allahabad-211004, INDIA (e-mail: dubeyhc@ieee.org). The research discussed in this report is financially supported by DRDO Linear Modulation Project at CRRAO Advanced Institute of Mathematics, Statistics and Computer Sciences, Hyderabad, INDIA. (PI-V.U. Reddy). The work was done at CRRAO AIMSCS during Summer 2011.
[2]Nandita received the Bachelor of Technology degree in Electronics and Communication Engineering from Birla Institute of Technology (BIT), Mesra-814142, INDIA (nandita16bit@gmail.com).
[3]Ashutosh Kumar Tiwari is research scholar in the Department of Electrical Engineering, MNNIT, Allahabad-211004, INDIA (email: aktiwari@ieee.org).

## II. MATHEMATICAL MODEL

In this section, mathematical expressions for various digitally modulated signals are presented followed by the expression for continuous wavelet transform (CWT) of these signals [13]. The received signal $r(t,i)$ is given by

$$r(t,i) = a_i e^{j*2\pi f*t} e^{j\theta} \sum_{k=1}^{N} s_k^i(t) g_T(t-(k-1)T) + n(t) \quad (1)$$

Where $s_k^i(t) = \tilde{s}(t) e^{j(2\pi f_c t + \theta_c)}$ (2)

is the transmitted digitally modulated signal and $\tilde{s}(t)$ is the corresponding complex baseband representation, $f$ is the carrier frequency, $\theta(t)$ is time-invariant carrier phase and $a_i$ is the flat fading coefficient of the channel. $n(t)$ being the complex AWGN noise introduced by the channel. The expression for various digitally modulated signals in their complex base-band representation is given as:

$$\tilde{s}_{QAM}(t) = \sum_{i=1}^{N}(A_i + j*B_i) g_T(t-iT) \quad (3)$$

$$\tilde{s}_{ASK}(t) = \sum_{i=1}^{N}(A_i) g_T(t-iT) \quad (4)$$

$$\tilde{s}_{PSK}(t) = \sqrt{S}\sum_{i=1}^{N} e^{j\Phi_i} g_T(t-iT) \quad (5)$$

$$\tilde{s}_{FSK}(t) = \sqrt{S}\sum_{i=1}^{N} e^{j(w_i t+\Phi_i)} g_T(t-iT) \quad (6)$$

The normalized versions of signal are given by

$$\bar{\tilde{s}}_{QAM}(t) = \sum_{i=1}^{N}(e^{\arctan(B_i/A_i)}) g_T(t-iT) \quad (7)$$

$$\bar{\tilde{s}}_{ASK}(t) = \sum_{i=1}^{N} sign(A_i) g_T(t-iT) \quad (8)$$

$$\bar{\tilde{s}}_{PSK}(t) = \sum_{i=1}^{N} e^{j\Phi_i} g_T(t-iT) \quad (9)$$

$$\bar{\tilde{s}}_{FSK}(t) = \sum_{i=1}^{N} e^{j(w_i t+\Phi_i)} g_T(t-iT) \quad (10)$$

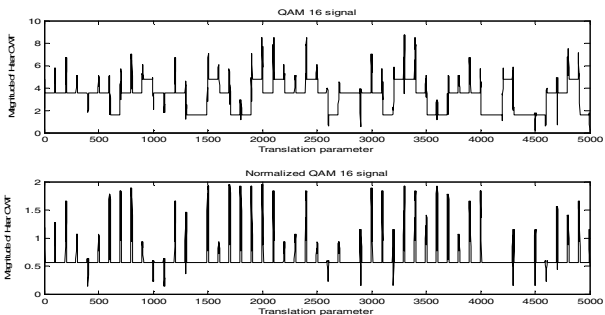

Fig.2. CWT of QAM-16 signal and its normalized form

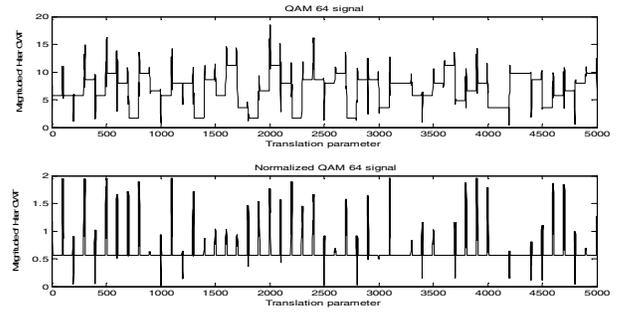

Fig.3. CWT of QAM-64 signal and its normalized form

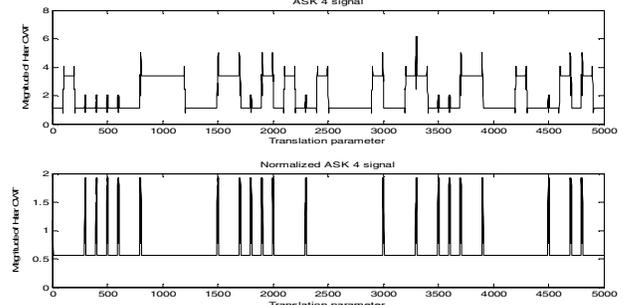

Fig.4. CWT of ASK-4 signal and its normalized form

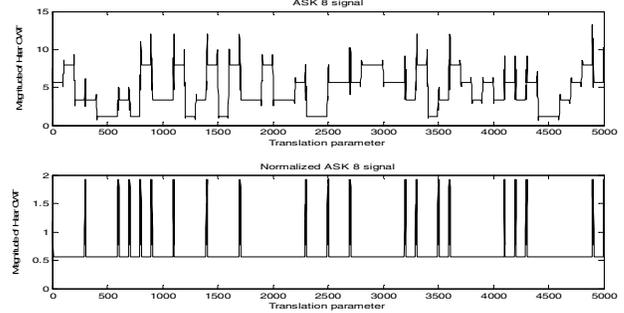

Fig.5. CWT of ASK-8 signal and its normalized form

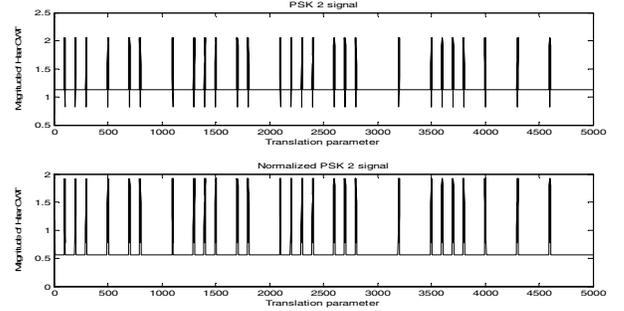

Fig.6. CWT of PSK-2 signal and its normalized form

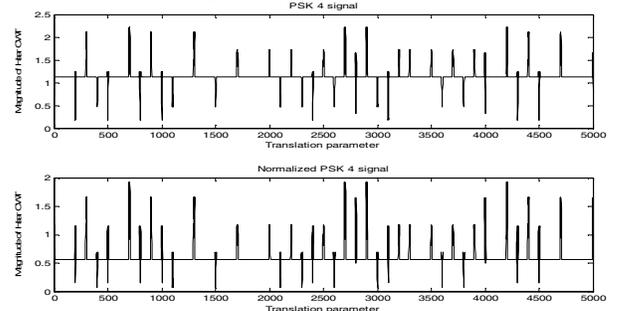

Fig.7. CWT of PSK-4 signal and its normalized form

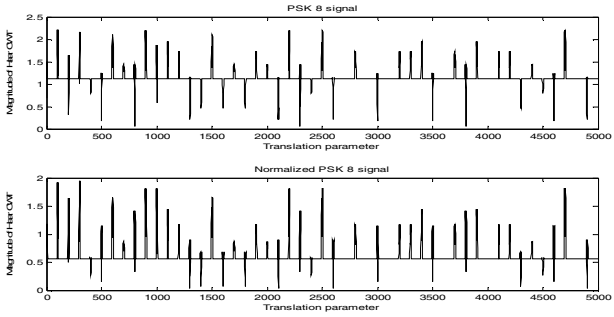

Fig.8. CWT of PSK-8 signal and its normalized form

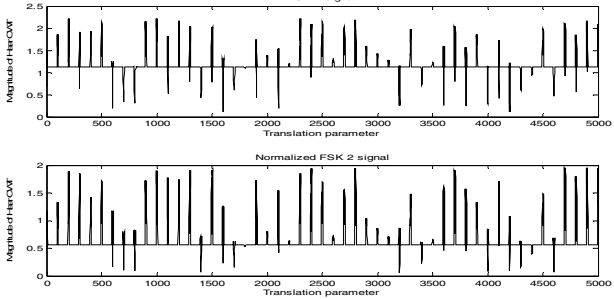

Fig.9. CWT of FSK-2 signal and its normalized form

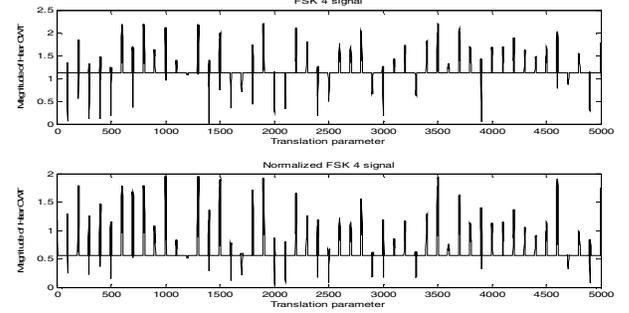

Fig.10. CWT of FSK-4 signal and its normalized form

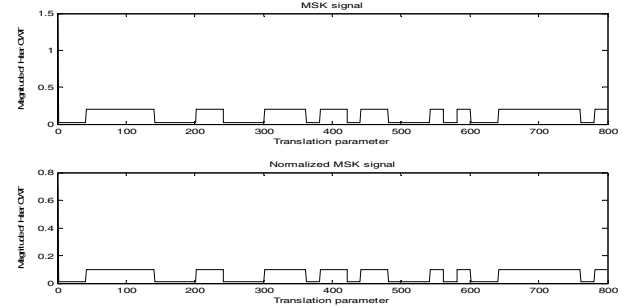

Fig.11. CWT of MSK signal and its normalized form

Following observation can be made from Fig.2 – Fig.11.
- The CWT of PSK is constant with as well as without normalization.
- The CWT of FSK is a multi-valued function with as well as without normalization.
- The CWT of ASK is a multi-valued function but CWT of normalized ASK is a constant function.
- The CWT of a QAM is a multi-valued function but CWT of normalized QAM is a constant function.
- The MSK is treated as special case of FSK where the phase varies continuously and the modulation index is 0.5 corresponding to minimum shift required for orthogonal signaling. So, the CWT of MSK as well as its normalized version is a two step function like BFSK.

Since the CWT of different signals have different variations, magnitude of CWT is suitable for extracting suitable features. Moments give information about underlying dynamics and hence moment of magnitude of CWT of signal as well as its normalized version contain signature of modulation type [14]. The higher order moment of the CWT of the received signal are the statistical features extracted from received signal. In this work, we have considered moments up to order 5. The moments of CWT are robust to the presence of noise so that they can be employed for modulation classification in presence of noise and disturbances as well. The normalized versions of the signals have the information about the dynamics of the signal which are otherwise absent in the CWT of the signal so the moments up to order 5 are calculated from both signal as well as its normalized version. The peaks occurring due to the phase changes at the instants where the wavelet accompanies a symbol change causes change in the statistics of the signal. So, the median filtering is employed for CWT of signals as well as CWT of the normalized signal. The length of the median filter used in this work is 5 [15]. The median filter removes the undesirable peaks and outliers occurring in the data.

## III. BLIND MODULATION CLASSIFICATION

### A. Principal Component Analysis (PCA)

PCA is a data analysis technique used for dimension reduction and feature extraction. It is a non-parametric method for extracting the useful dynamics from a noise corrupted data. A very comprehensive summary of PCA and related topics is given in [16]. In most of the practical scenario, we generally don't have any knowledge about dynamics of the systems but have only the measurements to deal with. PCA computes the transformed data which is the projection of the given data on most meaningful basis. The transformed data is able to reveal hidden dynamics in a more accurate way than the original data set. PCA reduces the dimension of data from *m* to *n* (*n<m*) while preserving the underlying dynamics. PCA finds a orthonormal transform *P* such that *PX=Y* where *X* is a matrix whose columns are original data, *Y* is a matrix whose columns are transformed data and *P* is a square matrix whose rows are the principal components. PCs are new set of basis vector for representing columns of *X*. PCA is done to tackle two inevitable nuisances, noise and redundancy in the measured data. PCA is based on second order statistics (SOS) and simple assumption of linearity on data set. The non-linear version of the PCA is called kernel-PCA. Kernel-PCA utilizes Mercer's kernel trick to first map the original non-linear data to kernel space where the data becomes linearly separable. This is followed by applying linear PCA in kernel space. Due to simplicity and good performance, linear PCA is used in this work. In context of dimension reduction of feature set obtained by CWT, *X* is *m* by *p* matrix where *m* is the number of features (in this case it is *20*) and *p* is the number of measurement trails (it is variable and can be *10, 20, 50,* and *100* depending on the classes considered for study). PCA finds an orthonormal matrix *P* and transforms the data to *Y* such that *Y=PX* and $C_Y = (1/[n-1])*Y*Y^T$ is diagonalized. There are two important ways to estimate PCs which are using covariance matrix or using singular value decomposition (SVD) [16]. In this work, covariance matrix was used to estimate PCs.

*B. Classification Sub-system*

This is the second sub-system of the pattern recognition based system for modulation classification. One versus all topology is used for deriving multi-class classifier from binary classifiers. All outputs are required to classify all 10 types of modulation types. Neural networks (NNs) are important machine learning tool which emulate the biological neural system. In this work, PNN have been proposed for classification of digitally modulated signals. MLPFN is used for performing a comparative study which is suggested in [3]. MLPFN typically consists of one input layer, one or more hidden layers and one output layer. The number of nodes in input layer is equal to the number of features in input data set and number of nodes in output layer is equal to number of classes. The number of hidden layers and number of nodes in each hidden layer is varied accordingly as par the need and specific requirement. With the increase in the hidden nodes or hidden layers, classification accuracy and generalization performance of the classifier is improved but at the same time computational burden increases. There is a tradeoff between computational burden and classification accuracy while dealing with such networks. MLPs have been popular in many classification tasks [17].

PNN has the advantage of much faster training over other NNs. Given enough input data, the PNN would converge to an optimum classifier. PNN allows incremental learning where the network can be redesigned without retraining the entire network and because of the statistical basis for the PNN, it can give an indication of the amount of evidence it has on the decision. PNN with Gaussian function has been used in this work [17]. The PNN consists of nodes allocated in three layers after the inputs and the structure is shown in Fig.12.

*Pattern Layer:* This layer assigns one node for each of the training pattern. There are two parameters associated with each pattern node. $w_i$ is the centre with the dimension *N by 1*, $\Sigma_i$ is the covariance matrix of *N by N* size, where *N* is the input vector length. The output of each of the pattern nodes is given as:

$$v_i = \exp\{-(\mathbf{x} - \mathbf{w_i})^T \Sigma_i^{-1} (\mathbf{x} - \mathbf{w_i})\}, i = 1, 2, \cdots, M \quad (22)$$

Where *M* is the number of input patterns.

*Summation Layer:* The number of nodes in this layer is the number of classes. Each of these nodes receives an input from each of the pattern nodes through a set of weights. The output of this layer is given as:

$$o_j = \sum_{k=1}^{M} \mathbf{w}_{jk}^{(s)} v_k, \ j = 1, 2, \cdots, L \quad (23)$$

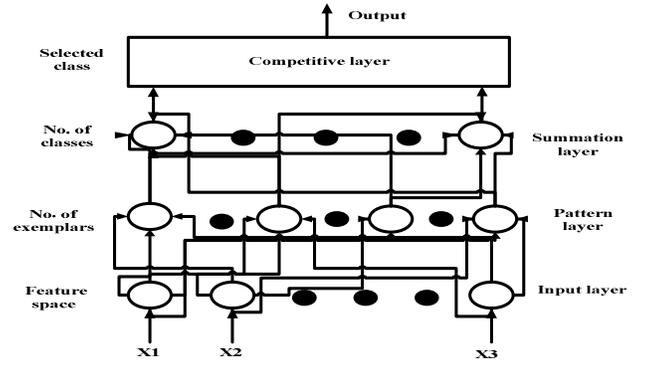

Fig.12. Probabilistic neural network

Where L is the number of classes. $\mathbf{w}_{jk}^{(s)}$ is the weight associated with the k-th pattern node to j-th summation node and $o_j$ is he output of the j-th summation node.

*Decision Layer:* This is the layer which indicates the particular class to which the present pattern belongs. A competitive procedure on the outputs of the summation layer decides the class to which the input patterns belong.

## IV. SIMULATION TESTS

This section summarizes the various case studies done. MATLAB have been used for carrying out case studies. The parameter of signals generated is given in Table I. The proposed technique has been tested with a flat fading channel with a phase offset taken randomly from range [-π:π]. Each type of digitally modulated signal is realized 10000 times. The 40 % of the data sets are utilized for training the network and the remaining 60 % are used for testing. The trained network is validated with respect to its classification accuracy. The validation is done by leave-one-out method which involves taking a data set out of training data set and using the rest of the training data for training and the taken out data set for testing. The 1 % has been taken as the validation limit.

TABLE I
PARAMETER OF SIGNALS USED IN STUDY

| Parameter | Value |
|---|---|
| Symbol rate | 100 |
| Number of samples per symbol | 100 |
| Carrier frequency | 100 kHz |
| Sampling frequency | 1 MHz |
| SNR | -2 dB to 8 dB |

TABLE II
SPECIFICATIONS OF MLPFN USED IN STUDY

| Parameter | Specification |
|---|---|
| No. of input nodes | 12 |
| No. of output nodes | 2 |
| No. of hidden layers | 2 |
| No. of nodes in first hidden layer | 10 |
| No. of nodes in second hidden layer | 15 |
| Learning algorithm | Resilient backpropagation |

TABLE III
PERFORMANCE OF PNN CLASSIFIER WITH SNR

| Performance | SNR (in dB) | | | | |
|---|---|---|---|---|---|
| | -2 | 1 | 5 | 6 | 8 |
| Training | 98.64 | 99.10 | 99.46 | 99.72 | 99.76 |
| Validation | 98.84 | 99.20 | 99.44 | 99.56 | 99.64 |
| Testing | 98.90 | 99.40 | 99.60 | 99.80 | 99.84 |
| Overall | 98.79 | 99.23 | 99.50 | 99.69 | 99.74 |

TABLE IV
PERFORMANCE OF MLPFN CLASSIFIER WITH SNR

| Performance | SNR (in dB) | | | | |
|---|---|---|---|---|---|
| | -2 | 1 | 5 | 6 | 8 |
| Training | 98.44 | 98.99 | 99.26 | 99.42 | 99.26 |
| Validation | 98.64 | 99.10 | 99.64 | 99.36 | 99.24 |
| Testing | 98.80 | 99.20 | 99.40 | 99.20 | 99.64 |
| Overall | 98.96 | 99.09 | 99.43 | 99.43 | 99.38 |

TABLE V
INTERCLASS PERFORMANCE OF PNN AT 2 dB SNR

| | PSK | FSK | QAM | ASK | MSK |
|---|---|---|---|---|---|
| PSK | 99.8 | 0 | 0.2 | 0 | 0 |
| FSK | 0 | 99.9 | 0 | 0 | 0.1 |
| QAM | 0 | 0 | 99.8 | 0.2 | 0 |
| ASK | 0 | 0 | 0.2 | 99.8 | 0 |
| MSK | 0 | 0.2 | 0 | 0 | 99.8 |

TABLE VI
INTERCLASS PERFORMANCE OF MLPFN AT 2 dB SNR

| | PSK | FSK | QAM | ASK | MSK |
|---|---|---|---|---|---|
| PSK | 99.6 | 0.1 | 0.3 | 0 | 0 |
| FSK | 0 | 99.8 | 0 | 0 | 0.2 |
| QAM | 0 | 0 | 99.6 | 0.4 | 0 |
| ASK | 0 | 0 | 0.4 | 99.6 | 0 |
| MSK | 0 | 0.4 | 0 | 0 | 99.6 |

TABLE VII
INTRACLASS PERFORMANCE OF PSK BY PNN AT 2 dB SNR

| | PSK-2 | PSK-4 | PSK-8 |
|---|---|---|---|
| PSK-2 | 100 | 0 | 0 |
| PSK-4 | 0.1 | 99.9 | 0 |
| PSK-8 | 0.1 | 0.1 | 99.8 |

TABLE VIII
INTRACLASS OF PSK BY MLPFN AT 2 dB SNR

| | PSK-2 | PSK-4 | PSK-8 |
|---|---|---|---|
| PSK-2 | 100 | 0 | 0 |
| PSK-4 | 0.2 | 99.8 | 0 |
| PSK-8 | 0.2 | 0 | 99.8 |

TABLE IX
PERFORMANCE OF PNN WITH MOTHER WAVELETS

| Mother wavelet | Overall Accuracy |
|---|---|
| Haar | 99.60 |
| Daubechies2 | 99.40 |
| Daubechies4 | 99.80 |
| Meyer | 99.20 |
| Morlet | 99.62 |

TABLE X
COMPARISON OF COMPUTATIONAL COMPLEXITY

| SNR (dB) | Training (MLP) | Testing (MLP) | Training (PNN) | Testing (PNN) |
|---|---|---|---|---|
| -2 | 2.63 | 0.62 | 0.49 | 0.27 |
| 2 | 2.33 | 0.56 | 0.42 | 0.25 |
| 4 | 2.43 | 0.52 | 0.44 | 0.24 |
| 6 | 2.21 | 0.48 | 0.46 | 0.23 |
| 8 | 2.12 | 0.42 | 0.48 | 0.22 |

The specifications of MLPFN used in this work are tabulated in Table II. Table III and IV gives the performance of PNN and MLPFN classifiers with different SNR taken as average on all cases. Table V and VI presents an instance of interclass recognition accuracy of both the classifiers at SNR of 2 dB. In addition, intra-class recognition of PSK signals is also given in Table VII and VIII. The performance of PNN for different mother wavelets of CWT is given in Table IX. Table X summarizes the comparison of computational complexity of both the algorithms. In generating Table X, 1000 separate data sets were used for both training and testing (for a binary classifiers corresponding to PSK2 class) on a portable computer using Intel (R) Core(TM) 2 Duo CPU T6500 with two clocks of frequency 2.10 GHz each and using MATLAB version 7.6.0 (R2008a) installed on Window 7 Home Basic operating system. Table XI and XII give the full class recognition of all studied cases for both PNN and MLPFN classifier respectively at 1 dB SNR. PNN is more accurate than MLPFN as results show.

## V. CONCLUSION

On the basis of simulation results we can conclude that PNN is a better classifier than MLPFN with CWT features both in terms of accuracy as well as computational complexity. Hence, PNN can be employed with CWT-PCA feature sub-set for accurate classification of modulation class of communication signals. Moreover, the investigation of various topologies of the ANN for the task is an interesting problem to deal with. One dimension of the research would be to study the robust of the scheme with time-varying channels, raised cosine pulse, and high noise levels. The second dimension can be investigation of other feature extraction techniques which can accurately track the signal transition and at the same time be robust to system changes.

## ACKNOWLEDGEMENT

Harish Chandra Dubey would like to thank Prof. V.U. Reddy (CRRAO AIMSCS Hyderabad) for helpful discussions.

TABLE XI
PERFORMANCE OF PNN AT 1 DB SNR

|  | QAM-16 | QAM-64 | ASK-4 | ASK-8 | PSK-2 | PSK-4 | PSK-8 | FSK-2 | FSK-4 | MSK |
|---|---|---|---|---|---|---|---|---|---|---|
| QAM-16 | 99.80 | 0.10 | 0.10 |  |  |  |  |  |  |  |
| QAM-64 | 0.05 | 99.90 |  | 0.05 |  |  |  |  |  |  |
| ASK-4 | 0.08 |  | 99.88 |  |  |  |  |  | 0.02 | 0.02 |
| ASK-8 |  | 0.06 |  | 99.94 |  |  |  |  |  |  |
| PSK-2 | 0.10 |  |  |  | 99.90 |  |  |  |  |  |
| PSK-4 | 0.05 |  |  |  | 0.05 | 99.90 |  |  |  |  |
| PSK-8 | 0.02 |  |  |  |  |  | 99.94 |  |  | 0.04 |
| FSK-2 |  |  |  |  |  |  |  | 99.98 |  | 0.02 |
| FSK-4 |  |  |  |  |  |  |  |  | 99.80 | 0.20 |
| MSK |  |  |  |  |  |  |  | 0.2 |  | 99.80 |

TABLE XII
PERFORMANCE OF MLPFN AT 1 DB SNR

|  | QAM-16 | QAM-64 | ASK-4 | ASK-8 | PSK-2 | PSK-4 | PSK-8 | FSK-2 | FSK-4 | MSK |
|---|---|---|---|---|---|---|---|---|---|---|
| QAM-16 | 99.60 | 0.20 | 0.20 |  |  |  |  |  |  |  |
| QAM-64 | 0.50 | 99.00 |  | 0.50 |  |  |  |  |  |  |
| ASK-4 | 0.60 |  | 99.20 |  |  |  |  |  | 0.1 | 0.1 |
| ASK-8 |  | 0.80 |  | 99.20 |  |  |  |  |  |  |
| PSK-2 | 0.60 |  |  |  | 99.40 |  |  |  |  |  |
| PSK-4 | 0.80 |  |  |  | 0.4 | 98.80 |  |  |  |  |
| PSK-8 | 0.4 |  |  |  |  |  | 99.00 |  |  | 0.6 |
| FSK-2 |  |  |  |  |  |  |  | 99.60 |  | 0.40 |
| FSK-4 |  |  |  |  |  |  |  |  | 99.40 | 0.60 |
| MSK |  |  |  |  |  |  |  | 0.40 |  | 99.60 |